\documentclass{article}

\usepackage[accepted]{icml2026}

\usepackage{microtype}
\usepackage{graphicx}
\usepackage{booktabs}
\usepackage{amsmath}
\usepackage{multirow}
\usepackage{url}
\usepackage{bm}
\microtypecontext{spacing=nonfrench}
\usepackage{xurl}
\usepackage[
    colorlinks=false,         
    citebordercolor={0 1 1},  
    linkbordercolor={1 0 0},  
    urlbordercolor={0 1 0}    
]{hyperref}

\usepackage{tikz}
\usetikzlibrary{shapes.geometric, arrows.meta, positioning, fit, backgrounds, calc, shadows}

\usepackage{algorithm}
\usepackage{algorithmic}

\icmltitlerunning{Tokenizer Transplantation for Bengali ASR}

\begin{document}

\twocolumn[
\icmltitle{Tokenizer Transplantation: Mitigating Autoregressive Collapse in Edge-Efficient Bengali ASR}

\begin{icmlauthorlist}
\icmlauthor{Sanjid Hasan}{kuet}
\icmlauthor{Md. Abdur Rahman}{mist}
\end{icmlauthorlist}

\icmlaffiliation{kuet}{Khulna University of Engineering \& Technology, Bangladesh}
\icmlaffiliation{mist}{Military Institute of Science and Technology, Bangladesh}

\icmlcorrespondingauthor{Sanjid Hasan}{hasan2203090@stud.kuet.ac.bd}

\icmlkeywords{ASR, Bengali, Tokenizer Fertility, Cross-Script, Adaptation}

\vskip 0.2in
]

\printAffiliationsAndNotice{}

\begin{abstract}
Lightweight speech recognition models are critical for edge deployment, yet highly optimized architectures like Moonshine often fail on morphologically rich, non-Latin languages such as Bengali. This study identifies the root cause of this failure as the model's English-centric byte-level tokenizer, which fragments Bengali words into high-fertility byte chains and triggers catastrophic autoregressive collapse during inference. To resolve this, a novel vocabulary transplantation pipeline is proposed to replace the decoder vocabulary with the native-script BanglaBERT WordPiece vocabulary and resize the corresponding token embedding matrix. Experimental results demonstrate a reduction in token fertility from 9.16 to 1.30. By decreasing autoregressive sequence length by 85.8\%, decoding instability is entirely mitigated. When evaluated on the 882-hour Lipi-Ghor dataset, the modified architecture achieves a competitive 21.54\% Word Error Rate (WER) and a Real-Time Factor (RTF) of 0.0053. Ultimately, this research provides a scalable, reproducible blueprint for cross-script adaptation of compact ASR models without the need for resource-intensive pre-training.
\end{abstract}
\section{Introduction}
Recent advancements in Automatic Speech Recognition (ASR) have achieved remarkable performance through large-scale, self-supervised learning. However, these models often rely on broad-coverage tokenizers designed for high-resource languages, which perform suboptimally on morphologically rich languages like Bengali. This mismatch results in high token fertility, leading to increased sequence length and autoregressive collapse during inference.

In this paper, we address this bottleneck by proposing a \textbf{Tokenizer Transplantation} pipeline. Rather than training from scratch, we surgically adapt a lightweight base architecture (e.g., Moonshine) by replacing its vocabulary with a native-script BanglaBERT tokenizer. We show that by decoupling acoustic representation learning from language modeling, we can stabilize the decoder and significantly improve performance on the 882-hour Lipi-Ghor dataset. Our approach provides a scalable blueprint for adapting efficient, pre-trained architectures to under-represented languages, ensuring that the benefits of ASR reach linguistically diverse communities. The full implementation details for our framework are available in our repository.\footnote{\url{https://github.com/Sanjidh090/moonshine-base-bn}}

\section{Related Work}

\subsection{Speech Representation Learning}
The current ASR landscape is dominated by frameworks such as \citet{baevski2020wav2vec} and \citet{babu2021xls}, which leverage self-supervised learning for robust feature extraction. While models like \citet{radford2023robust} have pushed the boundaries of accuracy via large-scale supervision, their computational footprint remains prohibitive for edge devices. Consequently, lightweight architectures like Moonshine \citep{usefulsensors2024moonshine, king2025flavors} have emerged as essential for local, resource-constrained deployment.

\subsection{Vocabulary Adaptation}
Tokenization quality is a primary determinant of model performance \citep{rust2021how}. Prior efforts, such as WECHSEL \citep{minixhofer2022wechsel} and FOCUS \citep{dobler2023focus}, have proposed innovative ways to initialize subword embeddings for cross-lingual transfer. However, these methods primarily address text-based Large Language Models (LLMs). Our work extends these principles to ASR by identifying "tokenizer transplantation" as a critical requirement for preventing autoregressive drift in morphologically rich scripts.

\subsection{Bengali ASR}
The development of Bengali ASR has been accelerated by datasets like \citet{alam2022bengali} and \citet{faisal2023ood}. While recent works  \citep{tabib2026bengaliloop}  have explored long-form ASR and speaker diarization \citep{bredin2020pyannote}, there remains a gap in adapting state-of-the-art decoders to these languages. By integrating techniques such as TokAlign \citep{liu2024tokalign} and Trans-Tokenization \citep{delobelle2024trans}, we provide a robust, reproducible methodology for language-specific vocabulary adaptation.
\section{Tokenizer Fertility and Decoding Instability}

The original Moonshine tokenizer was optimized primarily for English text. As a result, Bengali words undergo aggressive byte-level decomposition during tokenization. Tokenizer fertility is defined as the ratio of generated tokens to the total number of words:

\begin{equation}
\Phi = \frac{\text{Total Tokens}}{\text{Total Words}}
\end{equation}

where lower values indicate more compact tokenization. The baseline tokenizer produces $\Phi_{baseline} = 9.16$. In contrast, the transplanted Bengali tokenizer achieves $\Phi_{transplant} = 1.30$.

The fertility gap substantially affects autoregressive decoding. Long token chains increase the probability of inference-time error accumulation \citep{sennrich2016neural}. Although teacher-forced optimization remains stable, autoregressive generation becomes increasingly fragile, explaining the observed discrepancy between training loss and inference quality.

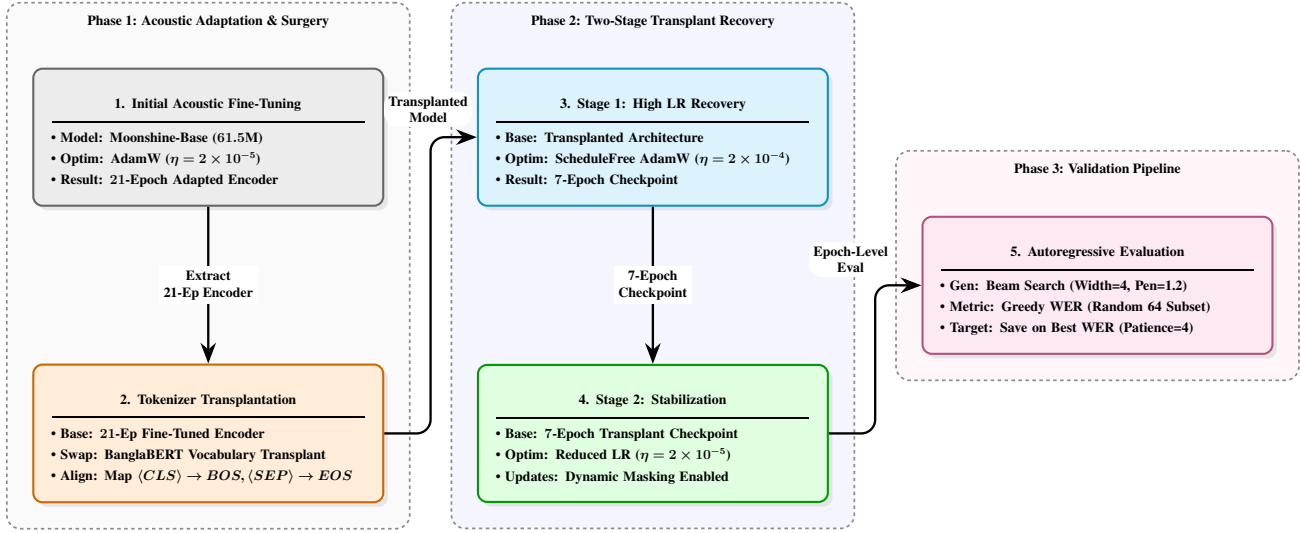
\begin{figure*}[t]
\centering
\resizebox{\textwidth}{!}{%
\begin{tikzpicture}[
    font=\large\bfseries,
    base_box/.style={
        rectangle,
        draw=black,
        line width=1.8pt,
        rounded corners=8pt,
        inner sep=16pt,
        text width=9.5cm,
        drop shadow={opacity=0.15, shadow xshift=4pt, shadow yshift=-4pt},
        align=left
    },
    arrow/.style={-{Stealth[scale=1.5]}, draw=black, line width=2pt},
    line/.style={draw=black, line width=2pt},
    group_bg/.style={
        rectangle,
        draw=black!45,
        dashed,
        line width=1.5pt,
        rounded corners=12pt,
        inner sep=22pt
    },
    arrow_label/.style={
        fill=white,
        inner sep=4pt,
        font=\large\bfseries,
        align=center
    }
]

\node[base_box, fill=gray!15, draw=gray!80!black] (A1) at (0, 4.5) {
    \begin{center}\textbf{1. Initial Acoustic Fine-Tuning}\end{center}
    \vspace{-0.2cm}\rule{\linewidth}{1.2pt}\vspace{0.1cm}\\
    \textbullet\ \textbf{Model:} \textbf{Moonshine-Base ($\bm{61.5}$M)}\\[0.8ex]
    \textbullet\ \textbf{Optim:} \textbf{AdamW ($\bm{\eta=2\times 10^{-5}}$)}\\[0.8ex]
    \textbullet\ \textbf{Result:} \textbf{$\bm{21}$-Epoch Adapted Encoder}
};

\node[base_box, fill=orange!15, draw=orange!80!black] (A2) at (0, -4.5) {
    \begin{center}\textbf{2. Tokenizer Transplantation}\end{center}
    \vspace{-0.2cm}\rule{\linewidth}{1.2pt}\vspace{0.1cm}\\
    \textbullet\ \textbf{Base:} \textbf{$\bm{21}$-Ep Fine-Tuned Encoder}\\[0.8ex]
    \textbullet\ \textbf{Swap:} \textbf{BanglaBERT Vocabulary Transplant}\\[0.8ex]
    \textbullet\ \textbf{Align:} \textbf{Map $\bm{\langle CLS \rangle \to BOS, \langle SEP \rangle \to EOS}$}
};

\node[base_box, fill=cyan!12, draw=cyan!70!black] (B1) at (13.5, 4.5) {
    \begin{center}\textbf{3. Stage 1: High LR Recovery}\end{center}
    \vspace{-0.2cm}\rule{\linewidth}{1.2pt}\vspace{0.1cm}\\
    \textbullet\ \textbf{Base:} \textbf{Transplanted Architecture}\\[0.8ex]
    \textbullet\ \textbf{Optim:} \textbf{ScheduleFree AdamW ($\bm{\eta=2\times 10^{-4}}$)}\\[0.8ex]
    \textbullet\ \textbf{Result:} \textbf{$\bm{7}$-Epoch Checkpoint}
};

\node[base_box, fill=green!12, draw=green!60!black] (B2) at (13.5, -4.5) {
    \begin{center}\textbf{4. Stage 2: Stabilization}\end{center}
    \vspace{-0.2cm}\rule{\linewidth}{1.2pt}\vspace{0.1cm}\\
    \textbullet\ \textbf{Base:} \textbf{$\bm{7}$-Epoch Transplant Checkpoint}\\[0.8ex]
    \textbullet\ \textbf{Optim:} \textbf{Reduced LR ($\bm{\eta=2\times 10^{-5}}$)}\\[0.8ex]
    \textbullet\ \textbf{Updates:} \textbf{Dynamic Masking Enabled}
};

\node[base_box, fill=magenta!10, draw=magenta!70!black] (C1) at (27, 0) {
    \begin{center}\textbf{5. Autoregressive Evaluation}\end{center}
    \vspace{-0.2cm}\rule{\linewidth}{1.2pt}\vspace{0.1cm}\\
    \textbullet\ \textbf{Gen:} \textbf{Beam Search (Width=4, Pen=1.2)}\\[0.8ex]
    \textbullet\ \textbf{Metric:} \textbf{Greedy WER (Random 64 Subset)}\\[0.8ex]
    \textbullet\ \textbf{Target:} \textbf{Save on Best WER (Patience=4)}
};

\begin{scope}[on background layer]
    \coordinate (A_top) at ($(A1.north) + (0, 1.2)$);
    \node[group_bg, fill=gray!4, fit=(A_top) (A1) (A2)] (bg_phase1) {};
    \node[anchor=north, font=\large\bfseries, text=black, yshift=-8pt] at (bg_phase1.north) {\textbf{Phase 1: Acoustic Adaptation \& Surgery}};

    \coordinate (B_top) at ($(B1.north) + (0, 1.2)$);
    \node[group_bg, fill=blue!4, fit=(B_top) (B1) (B2)] (bg_phase2) {};
    \node[anchor=north, font=\large\bfseries, text=black, yshift=-8pt] at (bg_phase2.north) {\textbf{Phase 2: Two-Stage Transplant Recovery}};

    \coordinate (C_top) at ($(C1.north) + (0, 1.2)$);
    \node[group_bg, fill=magenta!4, fit=(C_top) (C1)] (bg_phase3) {};
    \node[anchor=north, font=\large\bfseries, text=black, yshift=-8pt] at (bg_phase3.north) {\textbf{Phase 3: Validation Pipeline}};
\end{scope}

\draw[arrow] (A1.south) -- (A2.north) node[midway, arrow_label] {\textbf{Extract}\\\textbf{21-Ep Encoder}};
\draw[arrow, rounded corners=12pt] 
(A2.east) -- (6.75, -4.5) -- (6.75, 4.5) -- (B1.west) 
node[pos=0.5, above, xshift=-22pt, yshift=8pt, arrow_label]
{\textbf{Transplanted}\\ \textbf{Model}};
\draw[arrow] (B1.south) -- (B2.north) node[midway, arrow_label] {\textbf{7-Epoch}\\\textbf{Checkpoint}};
\draw[arrow, rounded corners=12pt] 
(B2.east) -- (20.25, -4.5) -- (20.25, 0) -- (C1.west) 
node[pos=0.5, above left, xshift=-4pt, yshift=6pt, arrow_label] 
{\textbf{Epoch-Level}\\\textbf{Eval}};
\end{tikzpicture}
}
\caption{Overview of the Tokenizer Transplantation Methodology.}
\label{fig:pipeline_diagram.pdf}
\end{figure*}

\section{Methodology}

To address the catastrophic decoding drift caused by high tokenizer fertility, a novel three-phase adaptation pipeline is proposed. Rather than training a model from scratch or replacing the vocabulary before fine-tuning, which often leads to catastrophic forgetting, the approach explicitly decouples acoustic representation learning from autoregressive language modeling. In this section, the sequential steps of the framework are detailed: initial acoustic adaptation of the base model, surgical transplantation of a native-script vocabulary, and a specialized recovery optimization schedule designed to stabilize the newly initialized decoder embeddings.

\subsection{Phase 1: Initial Acoustic Fine-Tuning}
Rather than replacing the tokenizer prior to any training which often destroys pretrained alignments and destabilizes the network our pipeline begins by fine-tuning the vanilla Moonshine-Base model on the Bengali dataset for 21 epochs. This allows the encoder to adapt its acoustic representations to Bengali phonetics. However, while the training loss converges, the decoder fails to produce coherent text due to the extreme sequence lengths generated by the native English-optimized tokenizer.

\subsection{Phase 2: Tokenizer Transplantation}

To resolve the decoding bottleneck while retaining the newly learned acoustic weights, tokenizer transplantation is performed on the 21-epoch fine-tuned model. The original decoder tokenizer is replaced with the BUET BanglaBERT WordPiece tokenizer \citep{bhattacharjee2022banglabert}. Decoder embeddings are mathematically resized from the original vocabulary size ($V_{old}$) to the Bengali target vocabulary ($V_{new}$).

Special tokens are remapped explicitly to preserve decoder compatibility (e.g., mapping BanglaBERT's $\langle CLS \rangle \rightarrow BOS$ and $\langle SEP \rangle \rightarrow EOS$). The 61.5M parameter encoder architecture remains unchanged, preserving the Bengali-adapted acoustic representations learned during Phase 1.

\subsection{Phase 3: Two-Stage Recovery Optimization}

Because the new decoder embeddings are randomly initialized while the acoustic encoder is already fine-tuned, standard optimization leads to transplant rejection and catastrophic forgetting. To stabilize decoder adaptation, a two-stage recovery fine-tuning approach is utilized, as illustrated in Figure \ref{fig:pipeline_diagram.pdf}:

\textbf{Stage 1 (High LR Recovery):} An aggressive initial learning rate ($\eta = 2 \times 10^{-4}$) is applied using the ScheduleFree AdamW optimizer \citep{defazio2024schedulefree}. This rapidly aligns the new textual embeddings to the acoustic latent space over 7 epochs without degrading the robust encoder weights.

\textbf{Stage 2 (Stabilization):} Training is resumed from the 7-epoch checkpoint with a heavily decayed learning rate ($\eta = 2 \times 10^{-5}$) to fine-tune the model for grammatical and phonetic accuracy.

To maximize memory efficiency, BF16 Automatic Mixed Precision (AMP), gradient checkpointing, and gradient accumulation (step 8) are applied to achieve an effective batch size of 32 throughout both stages.

\section{Experimental Setup}

All model adaptations and recovery fine-tuning schedules were implemented using PyTorch, leveraging a local \textbf{NVIDIA GeForce RTX 4070} hardware accelerator with \textbf{12.9 GB VRAM} to support memory-efficient execution via Automatic Mixed Precision (AMP) and gradient checkpointing. Transcript evaluations, Word Error Rate (WER) metric verification computations, and Real-Time Factor (RTF) speed benchmarking profiles were executed using the cloud-hosted \textbf{Kaggle} environment.

\subsection{Dataset}

Experiments utilize the Lipi-Ghor-882 dataset \citep{hasan2026make}, an 882-hour multi-speaker Bengali corpus curated from diverse open-source media. Audio is normalized to 16kHz mono. Silent segments and non-speech artifacts are aggressively filtered using Voice Activity Detection (VAD) via Pyannote \citep{bredin2020pyannote}. This builds upon foundational Bengali speech corpora such as OOD-Speech \citep{faisal2023ood} and Bengali Common Voice \citep{alam2022bengali} by offering long-form, highly diverse multi-speaker data.

\subsection{Evaluation Metrics}
Transcription quality is evaluated using Word Error Rate (WER) and Character Error Rate (CER), while inference efficiency is measured using Real-Time Factor (RTF). During training, Greedy WER is monitored on a randomized 64-sample validation subset at the end of each epoch, utilizing beam search (width = 4, repetition penalty = 1.2). Early stopping is triggered based on the best WER, with a patience of four epochs.

\section{Results}

\subsection{Mitigation of Autoregressive Collapse}
Tokenizer transplantation dramatically reduces Bengali sequence fragmentation (Table \ref{tab:fertility}). By compressing the token representations, autoregressive decoding becomes substantially more stable, completely preventing the catastrophic drift observed in the baseline.

\begin{table}[htbp]
\caption{Tokenizer Fertility Comparison}
\label{tab:fertility}
\centering
\vspace{0.1cm}
\begin{tabular}{lc}
\toprule
Tokenizer & Fertility ($\Phi$) \\
\midrule
Original Moonshine & 9.16 \\
Transplanted Bengali (BUET) & \textbf{1.30} \\
\bottomrule
\end{tabular}
\end{table}

\subsection{End-to-End ASR Performance}

To evaluate the real-world efficacy of the tokenizer transplantation pipeline, the proposed model is compared against a spectrum of zero-shot and fine-tuned baselines on the 5\% held-out Lipi-Ghor test set (comprising 5,931 utterances). 

As detailed in Table \ref{tab:performance}, the transplanted Moonshine-Base architecture achieves a highly competitive Word Error Rate (WER) of 21.54\%, substantially outperforming both the Zero-Shot Meta MMS 1B and the Fine-Tuned Conformer baselines. 

Most notably, the proposed model achieves a Character Error Rate (CER) of 10.79\%, the lowest among all tested architectures. This indicates that the native-script vocabulary empowers the decoder to construct morphologically complex Bengali words with higher sub-word accuracy than the byte-fallback approach utilized by Whisper. Furthermore, this accuracy is achieved with an order-of-magnitude reduction in parameters, matching the performance of the $\sim$769M parameter Faster Whisper Medium model with only $\sim$61.5M parameters.

\begin{table*}[htbp]
\caption{Global ASR Performance Comparison on the Lipi-Ghor Test Set. The proposed transplanted Moonshine model achieves a competitive WER and the lowest CER while operating with significantly fewer parameters than the Whisper and Conformer baselines.}
\label{tab:performance}
\vskip 0.15in
\begin{center}
\begin{small}
\resizebox{0.95\textwidth}{!}{
\begin{tabular}{llccc}
\toprule
\textbf{Pre-training Strategy} & \textbf{Model Architecture} & \textbf{Parameters} & \textbf{WER (\%)} & \textbf{CER (\%)} \\
\midrule
Zero-Shot & Seamless M4T-v2 & $\sim$2.3B & 66.71 & 45.54 \\
Zero-Shot & OpenAI Whisper large-v3 & $\sim$1.55B & 84.53 & 72.92 \\
Zero-Shot & Meta MMS 1B & $\sim$1B & 43.06 & 21.16 \\
Zero-Shot & Hishab TITU Conformer Large & $\sim$120M & 30.51 & 18.23 \\
\midrule
Fine-Tuned & Conformer Baseline (fast-conformer) & $\sim$120M & 24.67 & 15.56 \\
Fine-Tuned & Faster Whisper Medium (Tustugi) & $\sim$769M & 21.28 & 11.18 \\
\midrule
\textbf{Proposed} & \textbf{Moonshine Base (Transplanted Tokenizer)} & \textbf{$\sim$61.5M} & \textbf{21.54} & \textbf{10.79} \\
\bottomrule
\end{tabular}
}
\end{small}
\end{center}
\vskip -0.1in
\end{table*}

\subsection{Inference Efficiency and Edge Viability}

\begin{figure}[htbp]
    \centering
    \includegraphics[width=0.48\textwidth]{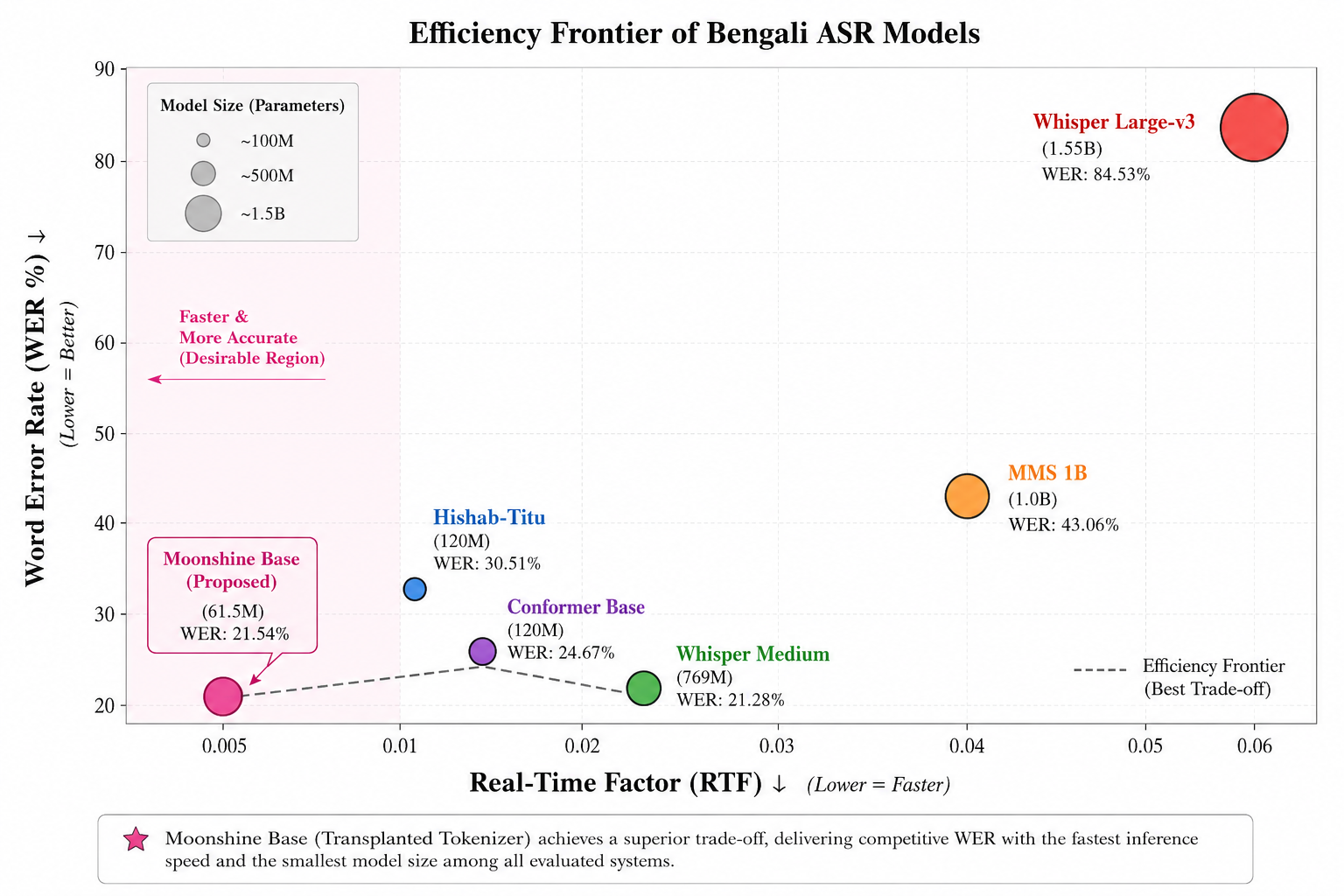}
    \caption{\textbf{Efficiency Frontier.} The proposed model (cyan) achieves a favorable trade-off between performance and efficiency.}
    \label{fig:efficiency_frontier}
\end{figure}

To contextualize the computational efficiency of the proposed architecture, inference speeds were compared against top-performing baselines evaluated during the DL Sprint 4.0 Bengali ASR competition \citep{hasan2026make}. Benchmarks are calculated based on the 22-hour hidden test set utilized in the competition. 

As demonstrated in Table \ref{tab:rtf}, while standard architectures like Whisper-Medium required heavily engineered, parallelized CTranslate2 pipelines \citep{klein2020opennmt} to achieve a Real-Time Factor (RTF) of $\sim$0.019, the proposed Moonshine-Base with the transplanted BanglaBERT tokenizer achieves an RTF of 0.0053 natively. 

Although the original vanilla Moonshine model processes the 22-hour set in slightly less time (RTF $\sim$0.0038), it yields catastrophic WERs near 100\% \citep{hasan2026make}. Tokenizer transplantation slightly increases sequence processing time due to accurate autoregressive generation, yet successfully salvages the model's accuracy while remaining 3.5$\times$ faster than highly optimized Whisper deployments and 2.2$\times$ faster than Conformer architectures.

\begin{table}[htbp]
\caption{Inference Efficiency on the 22-Hour DL Sprint Benchmark. The proposed model drastically outperforms competition baselines in speed while recovering accuracy.}
\label{tab:rtf}
\centering
\vspace{0.2cm}
\resizebox{\columnwidth}{!}{%
\begin{tabular}{llcc}
\toprule
\textbf{Model Architecture} & \textbf{Optimization} & \textbf{Time (22h)} & \textbf{RTF} \\
\midrule
Whisper-Medium & CTranslate2 / Dual T4 & $\sim$26 min & 0.0190 \\
Hishab-Titu Conformer & Standard PyTorch & $\sim$17 min & 0.0120 \\
\textbf{Proposed Moonshine-base} & \textbf{Tokenizer Transplant} & \textbf{$\sim$7 min} & \textbf{0.0053} \\
Moonshine-tiny & Baseline (Failed WER) & $\sim$5 min & 0.0038 \\
\bottomrule
\end{tabular}%
}
\end{table}

\section{Conclusion}
This study introduced tokenizer transplantation for the Bengali adaptation of the Moonshine ASR model. Analysis identified excessive tokenizer fertility as a primary source of autoregressive instability in compact models. By initially fine-tuning the acoustic representations, replacing the native tokenizer with a morphologically aligned Bengali tokenizer, and implementing a two-stage recovery training pipeline, inference stability on the Lipi-Ghor dataset was substantially improved. These findings demonstrate that decoupling acoustic adaptation from vocabulary alignment serves as a critical, underexplored strategy for democratizing lightweight ASR systems in low-resource languages.

\section*{Acknowledgments}
The authors express their sincere gratitude to the Department of Computer Science and Engineering (CSE) at Khulna University of Engineering \& Technology (KUET) for providing access to the specialized PC environment used to execute the fine-tuning and core adaptation stages of this research.


\end{document}